# OptMetaOpenFOAM: Large Language Model Driven Chain of Thought for Sensitivity Analysis and Parameter Optimization based on CFD

Yuxuan Chen[a], Long Zhang[a], Xu Zhu[a], Hua Zhou[a], Zhuyin Ren[a*]

[a] *Institute for Aero Engine, Tsinghua University, Beijing 100084, China*
\* *Corresponding author:* zhuyinren@tsinghua.edu.cn

**Abstract**

Merging natural language interfaces with computational fluid dynamics (CFD) workflows presents transformative opportunities for both industry and research. In this study, we introduce OptMetaOpenFOAM—a novel framework that bridges MetaOpenFOAM with external analysis and optimization tool libraries through a large language model (LLM)-driven chain-of-thought (COT) methodology. By automating complex CFD tasks via natural language inputs, the framework empowers non-expert users to perform sensitivity analyses and parameter optimizations with markedly improved efficiency. The test dataset comprises 11 distinct CFD analysis or optimization tasks, including a baseline simulation task derived from an OpenFOAM tutorial covering fluid dynamics, combustion, and heat transfer. Results confirm that OptMetaOpenFOAM can accurately interpret user requirements expressed in natural language and effectively invoke external tool libraries alongside MetaOpenFOAM to complete the tasks. Furthermore, validation on a non-OpenFOAM tutorial case—namely, a hydrogen combustion chamber—demonstrates that a mere 200-character natural language input can trigger a sequence of simulation, postprocessing, analysis, and optimization tasks spanning over 2,000 lines of code. These findings underscore the transformative potential of LLM-driven COT methodologies in linking external tool for advanced analysis and optimization, positioning OptMetaOpenFOAM as an effective tool that streamlines CFD simulations and enhances their convenience and efficiency for both industrial and research applications. Code is available at https://github.com/Terry-cyx/MetaOpenFOAM

## 1 Introduction

In recent years, with the development of Large Language Models (LLMs) [1-8], many fields have undergone significant transformations, with computer simulation software for optimization and analysis emerging as one of the most impacted domains. Traditionally, interactions with analysis and optimization software have been facilitated either through coding [9] or via graphical user interfaces (GUIs) [10, 11]. However, with advancements in natural language processing, integrating natural language into simulation-based analysis and optimization processes has emerged as a promising new approach [12, 13].

Computational Fluid Dynamics (CFD) is a computational technique that employs numerical methods and physical models to solve fluid flow, heat transfer, chemical reactions, and other related processes [14]. It is widely applied in various fields, including aerospace, energy [15-18], and biology. A complete CFD workflow typically includes modules such as Computer-aided Design (CAD) geometry generation, mesh generation, model and numerical parameter selection, solver execution, and post-processing. For beginners or researchers from other domains, any of these modules can be highly challenging. To lower the entry barriers for each stage in the workflow, frameworks using natural language as input have been proposed for tasks like CAD geometry generation [19], CFD simulation execution [12], and CFD post-processing [13].

Beyond these basic processes of CFD, subsequent analyses based on CFD simulations are also critical functionalities of industrial simulation software. Common analytical methods include Proper Orthogonal Decomposition (POD) [20], Dynamic Mode Decomposition (DMD) [21], and Chemical Explosion Mode Analysis (CEMA) [22], which focus on analyzing individual CFD simulation results. And for analyses such parameter sensitivity analysis, parameter calibration, and parameter optimization, performing multiple CFD simulations is essential. The former can naturally be seen as a specific type of CFD post-processing task, while the latter demands the creation of new frameworks capable of performing multi-simulation analysis using natural language inputs. This would further reduce the usability threshold for industrial CFD software.

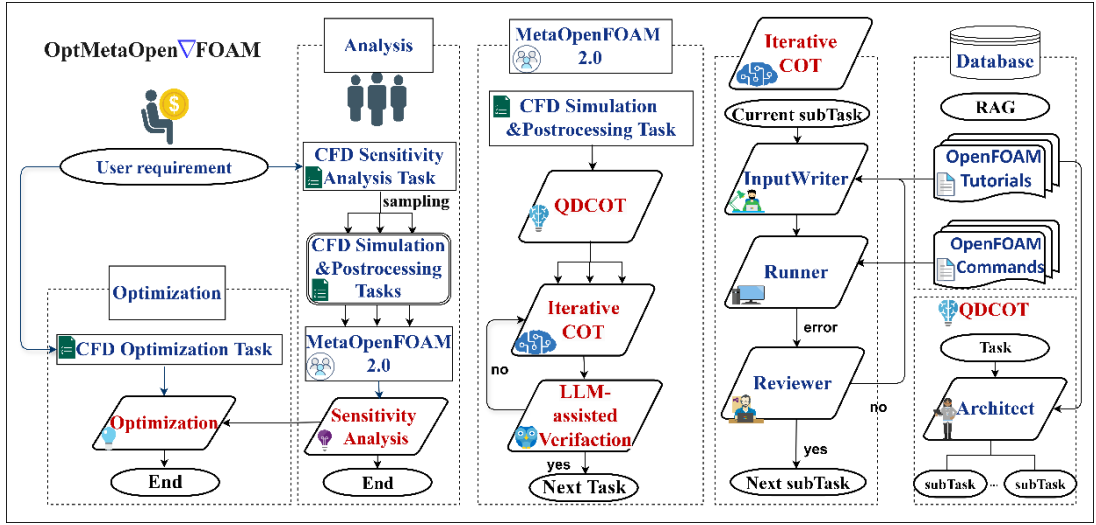

Figure 1: Framework of OptMetaOpenFOAM. Where QDCOT means chain of thought (COT) with question decomposition, Iterative COT (ICOT) means COT with iterative verification and refinement, and RAG means retrieval-augmented generation.

In recent years, the rapid development of large language models (LLMs), initially designed to tackle simple, quick-thinking tasks, has been exemplified by models such as GPT-4o [2] and DeepSeek v3 [6]. Later, models like GPT-o1 [4] and DeepSeek R1 [7] emerged to tackle more complex, slow-thinking problems. The former reflects pretrained scaling laws [1], while the latter demonstrates post-training scaling laws [8]. For CFD problems with natural language inputs, due to their complexity, high level of expertise, and reliance on external tools, using LLMs with general Chain of Thought (COT) [23-26] reasoning models may not be sufficient to accomplish the corresponding CFD tasks. Moreover, since CFD tasks typically follow a fixed workflow, it becomes feasible to develop a dedicated COT framework tailored specifically to CFD problems.

MetaOpenFOAM 1.0 [12] and 2.0 [13] have developed distinct COT structures to complete CFD simulation and postprocessing tasks, respectively. The proposed scaling laws demonstrated that increasing the number of COT steps improves the accuracy of the framework while increasing token usage, aligning with post-training scaling laws for LLM. Similarly, for analysis and optimization tasks that require multiple CFD simulations and postprocessing steps, specialized COT structures are necessary to facilitate the use of external tools (e.g., active subspace analysis [27], parameter optimization [28], etc.).

The structure of the paper is organized as follows: First, we introduce the basic framework of OptMetaOpenFOAM and the theoretical knowledge of the external tool libraries. Then, we describe the LLM settings used in the framework. Next, we present the input, thought process, and output results for specific cases in OptMetaOpenFOAM. Finally, we conclude with a summary.

First, OptMetaOpenFOAM establishes the corresponding CFD simulation, CFD postprocessing, CFD sensitivity analysis, and CFD parameter optimization tasks according to user requirements. It is important to note that the parameter optimization task is typically conducted based on the outcomes of the sensitivity analysis.

## 2 Methodology

### 2.1 MetaOpenFOAM 2.0 Framework

Figure 1 illustrates how OptMetaOpenFOAM leverages a chain-of-thought mechanism to process sensitivity analysis and parameter optimization tasks based on CFD, using natural language inputs.

Next, the framework extracts the sampling lower and upper range from the natural language input to perform sampling. If the sampling range is not specified, the LLM generates it automatically. A series of sampling points are then combined with the original CFD simulation and postprocessing tasks and fed into a natural language-driven CFD solver (i.e., MetaOpenFOAM 2.0 [13]). MetaOpenFOAM 2.0 primarily handles the CFD simulation and postprocessing tasks through Iterative COT and Question Decomposition COT (QDCOT) mechanisms.

After executing these CFD simulation and postprocessing tasks, a set of postprocessing outputs corresponding to the sampling points is obtained. An external sensitivity analysis tool is then invoked to complete the sensitivity analysis task via graphical visualizations and textual explanations. Finally, based on the response surface generated by the sensitivity analysis tool and the optimization target extracted from the user's requirements, an optimization function is executed, ultimately returning the optimized values of the input variables mentioned in user requirements.

## 2.2 External Tool Library

*Univariate Analysis and Optimization Tool Library*

In practical CFD simulations, it is often necessary to analyze the relationship between an independent variable and a dependent variable, calibrate parameters based on known experimental results, and adjust boundary or geometric parameters to achieve a specific target value. To accomplish these tasks with natural language inputs, corresponding external tool libraries must be integrated.

For univariate sensitivity analysis, the simplest method of graphical representation is used, followed by analysis performed by the LLM. In terms of univariate optimization, the L-BFGS-B optimization algorithm [29] is employed to solve problems with simple box constraints. This algorithm is a variant of the BFGS (Broyden-Fletcher-Goldfarb-Shanno) method [30], utilizing limited memory to approximate the Hessian matrix. It is particularly effective for large-scale optimization problems with box constraints, making it well-suited for the calibration and optimization tasks in CFD simulations. Given the constrained optimization problem:

$$\min f(x) \tag{1}$$

subject to the bounds:

$$l \leq x \leq u \tag{2}$$

Where $f(x)$ is the objective function, $x$ is the vector of decision variables, and $l$ and $u$ are the vectors of lower and upper bounds, respectively. The L-BFGS-B algorithm is shown in Algorithm 1.

---
**Algorithm 1** L-BFGS-B Algorithm
**Input:** Initial guess $\mathbf{x}_0$, tolerance $\varepsilon$, lower bounds $\mathbf{l}$, upper bounds $\mathbf{u}$
**Output:** Approximate solution $\mathbf{x}^*$
1: **function** L-BFGS-B($\mathbf{x}_0, \varepsilon, \mathbf{l}, \mathbf{u}$)
2:    $k \leftarrow 0$
3:    Initialize Hessian approximation $B_0$ (e.g., as the identity matrix) and memory for limited updates.
4:    **while** $\|\nabla f(\mathbf{x}_k)\|_2 \geq \varepsilon$ **do**
5:       Compute the gradient: $g_k \leftarrow \nabla f(\mathbf{x}_k)$
6:       Compute the search direction: $\mathbf{d}_k \leftarrow -B_k^{-1} g_k$
7:       **Project** $\mathbf{d}_k$ so that $\mathbf{l} \leq \mathbf{x}_k + \mathbf{d}_k \leq \mathbf{u}$
8:       Determine step length $\alpha_k$ via line search.
9:       Update the iterate: $\mathbf{x}_{k+1} \leftarrow \mathbf{x}_k + \alpha_k \mathbf{d}_k$
10:      Update Hessian approximation $B_k$ using the limited-memory BFGS formula.
11:      $k \leftarrow k + 1$
12:    **end while**
13:    **return** $\mathbf{x}_k$
14: **end function**

---

*Multivariable Analysis and Optimization Tool Library*

In practical CFD simulations, when the parameters to be analyzed or optimized increase in number, the complexity of the analysis and optimization paths also significantly increases.

In terms of analysis, one key concern is the sensitivity analysis of multiple parameters, as well as the analysis of their controlling mechanisms and uncertainties. In this study, the active subspace method [27, 28] is employed to handle multivariable sensitivity analysis, controlling mechanism identification, and uncertainty analysis tasks. This method effectively identifies dominant directions in high-dimensional parameter spaces, enabling a more efficient exploration of the parameter space and a clearer understanding of the underlying mechanisms and uncertainties.

Given uncertain inputs $x \in \mathbb{R}^m$, where $f$ maps the inputs, the inputs are normalized as:

$$x_i = \frac{x_i - x_L}{x_U - x_L} \tag{3}$$

The covariance matrix $C$ is obtained by eigenvalue decomposition:

$$C = \int \nabla_x f(x) \nabla_x f(x)^T \pi(x) dx \tag{4}$$
$$= W \Lambda W^T$$

where $\pi(x)$ is probability distribution of $x$, and the gradient is:

**Algorithm 2** Active Subspace-based Optimization Procedure

**Input:** Dimension $m$, oversampling factor $\theta \in [2, 10]$, simulation model $f(\mathbf{x})$, lower and upper bounds for sampling $lb, ub$
**Output:** Summary plot for 1-D structure of $f(\mathbf{x})$ and optimal solution $\mathbf{x}^*$
1: **function** AS_OPTIMIZATION($m, \theta, f, lb, ub$)
2:    **(1) Sampling and Normalization:**
      Set $N \leftarrow \theta m$
      Draw $N$ i.i.d. samples $\{\mathbf{x}^{(j)}\}_{j=1}^{N}$ between $lb$ and $ub$ using uniform sampling method
      Normalize each sampled element so that $x_j \in [0, 1]$
3:    **(2) Model Evaluation:**
      Run simulations for all $N$ cases to obtain the quantities of interest:
      $\mathbf{Q} = [Q_1, \ldots, Q_N]^T$
4:    **(3) OLS Regression:**
      Estimate $\mathbf{b} = [b_1, \ldots, b_m]^T$ via an ordinary least squares (OLS) model as an example:
      $\hat{Q} = c + \mathbf{x}^T \hat{\mathbf{b}}$
5:    **(4) Identify 1-D Subspace:**
      Compute $\hat{\mathbf{w}} \leftarrow \mathbf{b}/\|\mathbf{b}\|$
      Generate a summary plot of $\mathbf{Q}$ vs. $\hat{\mathbf{w}}^T \mathbf{x}$ to verify the 1-D structure
6:    **(5) Error/Confidence Analysis:**
      Use bootstrap to estimate the error bound, confirming the validity of the 1-D structure
7:    **(6) Construct Reduced Model:**
      Define $z = \hat{\mathbf{w}}^T \mathbf{x}$, and consider $g(z) \approx f(\mathbf{x})$
8:    **(7) Optimization:**
      Solve $\min_z g(z)$
      Map $z^*$ back to $\mathbf{x}^*$
9:    **(8) Post-Processing:**
      Validate the optimized solution $\mathbf{x}^*$ with the original high-dimensional model $f(\mathbf{x})$
10:   Return $\mathbf{x}^*$, summary plots, and error estimates
11: **end function**

$$\nabla_x f(\mathbf{x}) = \begin{bmatrix} \frac{\partial f}{\partial x_1}(\mathbf{x}) \\ \vdots \\ \frac{\partial f}{\partial x_m}(\mathbf{x}) \end{bmatrix} \qquad (5)$$

The eigenvalue matrix $\mathbf{\Lambda} = \text{diag}(\lambda_1, \cdots, \lambda_m)$ with eigenvalues $\lambda_1 \geq \cdots \geq \lambda_m \geq 0$ is defined, and the eigenvectors $\mathbf{w}_j$ satisfy:

$$\begin{aligned}\lambda_i &= \mathbf{w}_i^T \mathbf{C} \mathbf{w}_i \\ &= \int (\nabla_x f(\mathbf{x})^T \mathbf{w}_i)^2 \pi(\mathbf{x}) d\mathbf{x}\end{aligned} \qquad (1)$$

This indicates that eigenvalue $\lambda_j$ can represent how much $f(\mathbf{x})$ will change when advancing $\mathbf{x}$ along the direction of corresponding $\mathbf{w}_j$. Therefore, in the already sorted sequence of eigenvalues, the difference between two eigenvalues can reflect the change of $f(\mathbf{x})$. For example, if $\lambda_n$ is much larger than $\lambda_{n+1}$, i.e., $\lambda_n \gg \lambda_{n+1} (n < m)$, the eigenvalue matrix and eigenvector matrix can be divided into two parts:

For large $\lambda_n$ relative to $\lambda_{n+1}$, the eigenvalue eigenvector matrices can be split:

$$\mathbf{\Lambda} = \begin{bmatrix} \mathbf{\Lambda}_1 & \\ & \mathbf{\Lambda}_2 \end{bmatrix}, \mathbf{W} = [\mathbf{W}_1 \quad \mathbf{W}_2] \qquad (7)$$

where $\mathbf{\Lambda}_1$ holds the dominant eigenvalues. The change in $f$ along the direction of $\mathbf{W}_1$ is much larger than along $\mathbf{W}_2$. If small eigenvalues are negligible, $\nabla_{\mathbf{W}_2^T \mathbf{x}} f(\mathbf{x}) = 0$, leading to the active subspace $\mathbf{AS} = [\mathbf{w}_1, \cdots \mathbf{w}_n]$. Thus, $f(\mathbf{x})$ can be approximated as $f(\mathbf{x}) \approx g(\mathbf{W}_1^T \mathbf{x}) = g(\mathbf{z})$, with $\mathbf{z} = \mathbf{AS}^T \mathbf{x}$. For high-dimensional problems, $\mathbf{C}$ is approximated via linear regression. Assuming $\nabla_x f(\mathbf{x}) \approx \mathbf{b}$ (OLS Regression), the active subspace is obtained by:

$$\hat{\mathbf{C}} = \int \mathbf{b}\mathbf{b}^T \pi(\mathbf{x}) d\mathbf{x} = \mathbf{b}\mathbf{b}^T = \hat{\mathbf{w}}\hat{\lambda}\hat{\mathbf{w}}^T \qquad (8)$$

while the $\hat{\lambda} = \|\mathbf{b}\|^2$, $\hat{\mathbf{w}} = \mathbf{b}/\|\mathbf{b}\|$, and the one-dimensional subspace is the active direction $\mathbf{AS} \equiv \hat{\mathbf{w}}$. The model $f$ is validated through a univariate relationship of each quantity of interest (QoI) against $\hat{\mathbf{w}}^T \mathbf{x}$, with $\hat{\mathbf{w}}$ indicating sensitivity coefficients in the input space. The specific multivariable analysis and optimization algorithm is shown in Algorithm 2.

In fact, in addition to OLS regression, the integrated tool library also selects other regression methods based on the fitting results, such as a global quadratic model (QPHD) regression [27].

## 3 Setup

MetaGPT v0.8.0 [31] was chosen for the integration of different agents, while OpenFOAM 10 [9] was employed for CFD simulations due to its stability and dependability as an open-source solver. GPT-4o [2] was selected as the primary LLM, owing to its exceptional performance. To minimize randomness in the generated output, the temperature parameter, which controls the degree of randomness in LLM-generated text, was configured to 0.01, ensuring more deterministic and focused results. The impact of temperature settings on model performance is discussed in [12].

In terms of RAG, LangChain v0.1.19 [32] facilitated the connection between the agents and the database. The FAISS vector store [33], recognized for its high efficiency and ease of use, was utilized as the vector store for the database, and OpenAIEmbeddings were chosen for embedding the data segments. The "similarity" approach was employed to identify and match related chunks of data. The simplest stacking approach was used, combining retrieved documents with user input messages. Additional information can be found in the code repository: https://github.com/Terry-cyx/MetaOpenFOAM. In the external analysis and optimization tool library, the univariate optimization method, the L-BFGS-B optimization algorithm, is invoked through the bounded method in scipy.optimize. For multivariable analysis and optimization, the active subspace method is used [27].

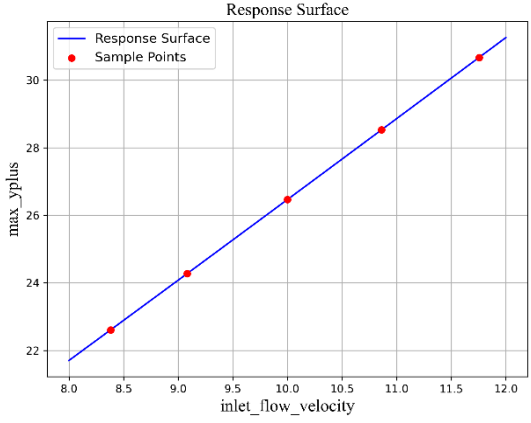

Figure 2: Response surface of the inlet flow velocity versus the max yPlus in PitzDaily.

## 4 Results

The evaluation metrics used previously were based on the CFD framework with natural language inputs, with a primary focus on the user requirements for Executability (ranging from 0 to 7, corresponding to failure to flawless), Cost (including token usage, number of iterations, running time, etc.), and Pass@k [34, 35]. These metrics remain significant in OptMetaOpenFOAM, but due to the integration of fixed interfaces after completing the CFD simulation and postprocessing tasks, the Executability, Pass@k, and number of iterations remain consistent with the previous statistics in MetaOpenFOAM 2.0 [13]. Regarding Cost, token usage slightly increases due to the addition of two new modules, while the running time increases linearly with the addition of CFD simulation tasks. Therefore, in addition to these three metrics, the evaluation of OptMetaOpenFOAM increasingly focuses on the result presentation.

In this section, we will analyze five cases: PitzDaily, CounterFlowFlame, BuoyancyCavity, HIT, and Hydrogen Combustion Chamber. Among these, the first four cases are OpenFOAM tutorials, while the latter is not. All the basic cases have been incorporated into the database required for the RAG technique. All the figures presented in this section were generated using OptMetaOpenFOAM based on user prompts. It is important to note that the prompt for textual response is provided in **Appendix** A, the full textual analysis provided by OptMetaOpenFOAM is included in **Appendix** B and the thought process provided by OptMetaOpenFOAM is included in **Appendix** C.

### 4.1 PitzDaily

This case is an incompressible flow, simulated using the RANS method. The following prompt is used to perform simulation for this case.

**CFD simulation task**: Do a RANS simulation of incompressible PitzDaily flow.

**CFD postprocessing task**: Extract max yplus at latest time through post-processing

Investigating the relationship between inlet velocity and y+ is one of the common CFD sensitivity analysis tasks. A similar analysis can be performed for the relationship between inlet velocity and the Courant number. The following prompt is used to study this case.

**CFD analysis tasks**: ① Analyze the effect of the inlet flow velocity on max yplus. ② Analyze the effect of the inlet flow velocity and inlet turbulent kinetic energy on max yplus.

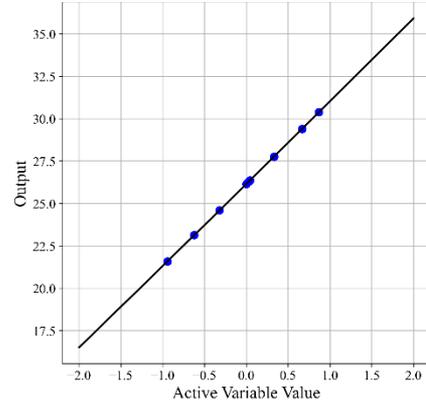
(a)

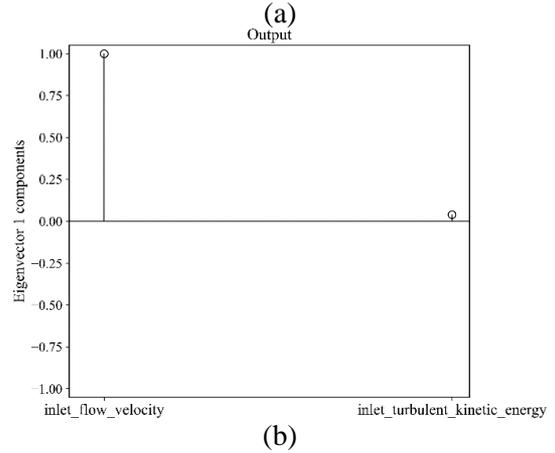
(b)

Figure 3: Effect of inlet flow velocity and inlet turbulent kinetic energy on max y+ in PitzDaily case. (a) response surface and (b) components of active direction ($\hat{w}$).

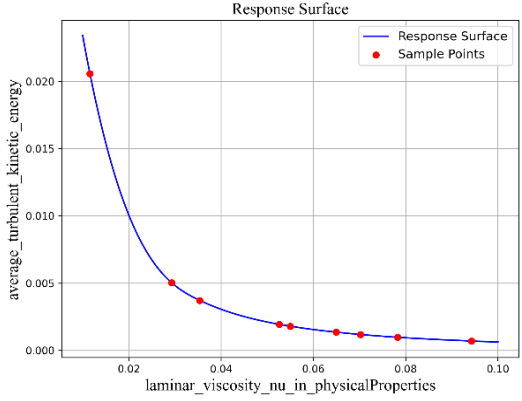

Figure 4: Effect of laminar viscosity on the average turbulent kinetic energy in HIT case.

**CFD optimization task**: Determine the optimal inlet flow velocity and inlet turbulent kinetic energy at which the max yplus should be as close to 25 as possible.

The response surface plot in Figure 2 shows a positive linear relationship between inlet flow velocity and max y+, with values increasing from 22 to 31 as velocity rises from 8 m/s to 12 m/s. This trend is consistent with theoretical expectations, indicating higher shear stress at the wall. As velocity increases, the boundary layer becomes thinner, requiring finer near-wall resolution. For low velocities, max y+ remains in the low-Reynolds number range, while higher velocities push the values into the logarithmic layer, suggesting the need for wall functions. Subsequently, parameter optimization was performed on the response surface. To ensure that the simulated max y+ remains below 25, the inlet flow velocity was adjusted to 9.39 m/s.

The response surface analysis in Figure 3 reveals a strong linear relationship between an active variable—combining inlet flow velocity and turbulent kinetic energy—and max y+, with simulation results closely matching model predictions. Eigenvector analysis indicates that inlet flow velocity (component near 1.0) is the dominant factor affecting max y+, while turbulent kinetic energy plays a negligible role.

### 4.2 HIT

Homogeneous isotropic turbulence (HIT) represents an idealized turbulent state in which the statistical properties of the flow remain invariant under any spatial translations or rotations, meaning there is no preferential direction or positional dependency. In this test case, the fluid exhibits no significant mean flow, and turbulence consists solely of random fluctuations. Consequently, HIT serves as an ideal benchmark for investigating the energy cascade process, validating Kolmogorov's 1941 theory, and testing various turbulence closure models. In numerical simulations, an initial velocity field is typically constructed under periodic boundary conditions to match a prescribed energy spectrum, while an external forcing term is employed to balance viscous dissipation and maintain a statistically steady state. Owing to its simple geometry and high degree of symmetry, the homogeneous isotropic turbulence test case not only helps reveal the mechanisms of energy transfer from large scales to small scales but also provides a unified benchmark for assessing different numerical methods and turbulence models. The following prompt is used to perform simulation for this case.

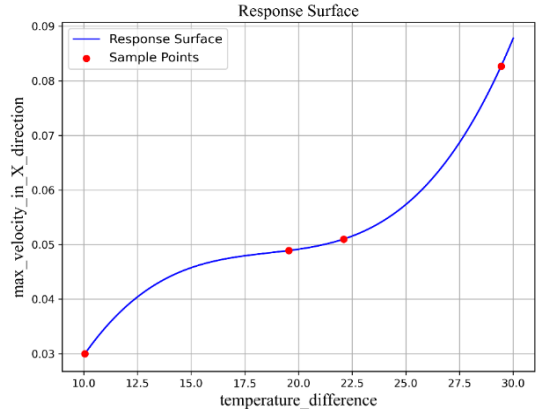

Figure 5: Effect of ΔT between two boundaries on maximum X velocity in BuoyantCavity.

**CFD simulation task**: Perform a DNS simulation of incompressible forced homogeneous isotropic turbulence.

**CFD postprocessing task**: Extract the average turbulent kinetic energy (average($1/2*U^2$)) at latest time through post-processing

**CFD analysis task**: Analyze the effect of the laminar viscosity nu_in_physicalProperties (from 0.01 to 0.1) on the average turbulent kinetic energy (average($1/2*U^2$)).

**CFD optimization task**: Determine the optimal nu_in_physicalProperties at which the average turbulent kinetic energy is near 0.01.

The response surface in Figure 4 reveals a strong inverse relationship between laminar viscosity ($\nu$) and average turbulent kinetic energy. As $\nu$ increases from 0.01 to 0.1, turbulent kinetic energy declines sharply—indicating that higher viscosity enhances energy dissipation and dampens turbulence. Subsequently, parameter calibration was performed on the response surface. Assuming the desired average turbulent kinetic energy is around 0.01, the viscosity parameter was adjusted to 0.02 m²/s.

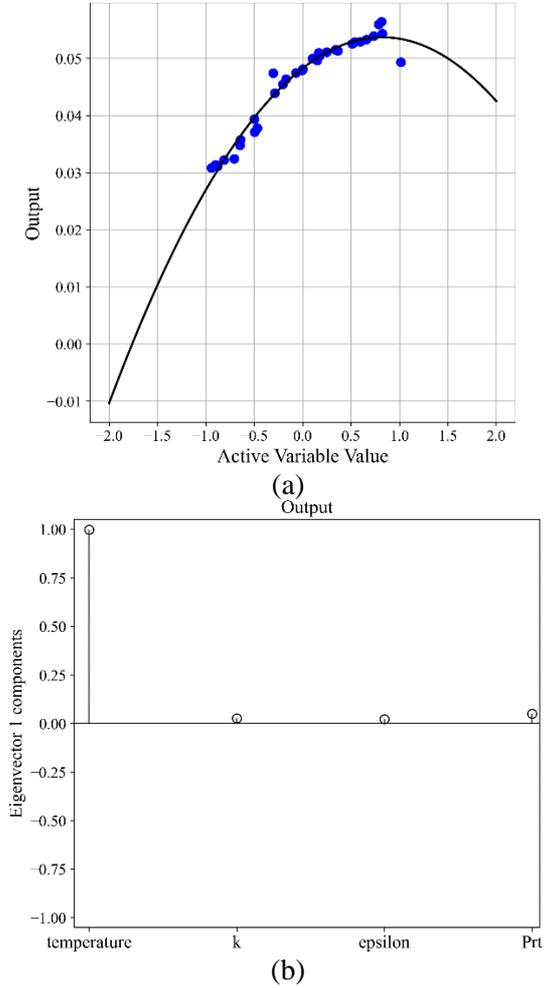

Figure 6: Effect of ΔT, k, epsilon and Pr$_t$ of boundary on maximum X velocity in BuoyantCavity. (a) response surface and (b) components of active direction ($\hat{w}$).

### 4.3 BuoyantCavity

The BuoyantCavity case represents a canonical benchmark for studying buoyancy-driven flows in a confined cavity. In this configuration, a fluid-filled cavity is subjected to differential heating, which induces natural convection as buoyancy forces overcome viscous damping. The resulting flow is characterized by the interplay between thermal gradients and momentum transport, leading to the development of recirculation zones and complex vortical structures. This case is widely used to validate numerical solvers for coupled heat and fluid flow problems, as it involves solving the incompressible Navier–Stokes equations augmented by an energy transport equation under the Boussinesq approximation. Owing to its moderate geometric complexity and well-defined boundary conditions, the BuoyantCavity test case serves as an effective benchmark to assess the accuracy, convergence, and stability of various discretization schemes and turbulence models in simulating natural convection phenomena. The following prompt is used to perform simulation for this case.

**CFD simulation task**: do a RANS simulation of buoyantCavity using buoyantFoam, which investigates natural convection in a heat cavity; the remaining patches are treated as adiabatic.

**CFD postprocessing task**: Extract the max velocity in X direction at latest time through post-processing.

**CFD analysis tasks**: ① Analyze the effect of the temperature difference betwwen the hot and cold (from 10 K to 30 K) on the max velocity in X direction. ② Analyze the effect of the temperature_difference_betwwen_hot_and_cold (from 10 K to 30 K), k_of_all_boundarys (from 1e-04 to 1e-03), epsilon_of_all_boundarys (from 1e-06 to 1e-05) and Prt_of_all_boundarys_in_alphat (from 0.6 to 1.0) on the max velocity in X direction.

**CFD optimization task**: Determine the optimal temperature difference betwwen the hot and cold at which max velocity in X direction is near 0.07 m/s in a simulation.

Response surface in Figure 5 shows that an increase in temperature difference (ΔT) between the hot and cold patches leads to a nonlinear rise in maximum velocity in the X direction (Umax,x). As ΔT increases from 10 K to 30 K, buoyancy effects enhance convection, accelerating the flow and potentially increasing turbulence. The relationship between ΔT and Umax,x is nonlinear. Subsequently, parameter calibration was conducted on the response surface. Assuming the maximum velocity in the X direction was measured to be 0.07 m/s, the temperature difference between the two walls was determined to be 27.7 K through the use of an external optimization module.

The Active Subspace Method analysis of a buoyant cavity simulation shown in Figure 6 reveals that the temperature difference between the hot and cold surfaces is the dominant factor influencing the maximum velocity in the X direction. The response surface shows a nonlinear relationship, with velocity increasing initially and then plateauing at higher temperature differences. The eigenvector analysis indicates that temperature difference has the strongest impact, while the turbulence parameters (k, epsilon, $Pr_t$) have minimal effect. For velocity control, adjusting the temperature difference is the most effective approach.

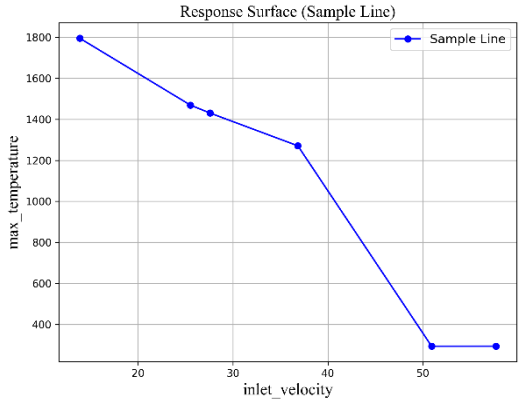

Figure 7: Effect of inlet velocity on maximum temperature in CounterFlowFlame2D.

### 4.4 CounterFlowFlame2D

CounterFlowFlame2D is a canonical two-dimensional counterflow flame configuration widely used in combustion simulations. In this setup, streams of fuel and oxidizer are injected from opposing inlets, and their interaction within the computational domain gives rise to a stabilized flame region. By solving the conservation equations for mass, momentum, energy, and chemical species under low Mach number assumptions—and incorporating detailed chemical kinetics—the case is capable of capturing flame stabilization and potential extinction phenomena. Owing to its simple geometry and well-defined boundary conditions, CounterFlowFlame2D serves as an ideal validation platform for evaluating reaction–flow coupling models, flame structure, and heat transfer mechanisms, while also providing a crucial benchmark for the study of flame dynamics and the convergence properties of numerical methods. The following prompt is used to perform simulation for this case.

**CFD simulation task**: do a 2D laminar simulation of counterflow flame using reactingFoam.

**CFD postprocessing task**: Extract the max temperature at latest time through post-processing.

**CFD analysis tasks:** ① Analyze the effect of inlet velocity (from 10.0 to 60.0 m/s) on max temperature. ② Analyze the effect of inlet velocity (from 10.0 to 60.0 m/s) and inlet temperature (from 243 to 343 K) on max temperature.

**CFD optimization tasks**: determine the optimal inlet velocity at which max temperature is blow 1000 K

The analysis shown in Figure 7 of the counterflow flame simulation shows an inverse relationship between inlet velocity and maximum temperature. As inlet velocity increases from 10 m/s to 60 m/s, the maximum temperature decreases, with a sharp drop occurring beyond 40 m/s, suggesting a flame extinction limit. At lower velocities, the flame remains stable, leading to higher temperatures, while higher velocities enhance convective cooling, reducing peak temperature. Subsequently, optimization was performed on the response surface. The objective of the optimization task was to determine the inlet velocity (or strain rate) at which flame quenching occurs. Ultimately, by using a simple prompt and invoking an external optimization module, the minimum quenching inlet velocity was found to be 41.2 m/s.

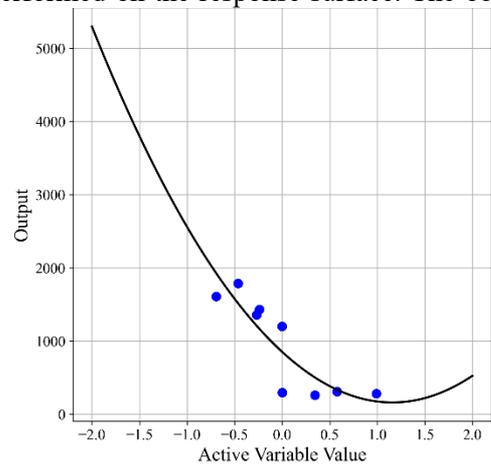

(a)

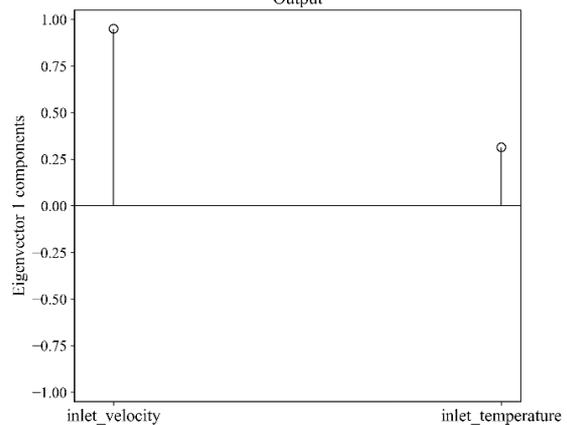

(b)

Figure 8: Effect of inlet flow velocity and inlet temperature on maximum temperature in CounterFlowFlame2D. (a) response surface and (b) components of active direction ($\hat{w}$).

The analysis shown in Figure 8 using the Active Subspace Method reveals that inlet velocity has a dominant influence on the maximum temperature in a laminar simulation. The response surface shows that as inlet velocity increases, the maximum temperature decreases, while higher inlet temperature results in higher maximum temperature. The eigenvector analysis further confirms that inlet velocity contributes significantly to the first eigenvector, indicating its primary role in controlling the maximum temperature, while inlet temperature has a negligible effect. Therefore, for simulations in a counterflow flame scenario, controlling inlet velocity will have a more significant impact on maximum temperature than inlet temperature.

### 4.5 Hydrogen Combustion Chamber

The Hydrogen Combustion Chamber case is a non‑OpenFOAM tutorial benchmark designed to analyze hydrogen flame propagation using natural language input. In this configuration, a two‑dimensional computational domain with fully periodic boundary conditions on all four sides is employed. The simulation uses the Launder‑Sharma low Reynolds number $k-\varepsilon$ two‑equation turbulence model [36] to capture the flame dynamics following ignition initiated at the lower left corner. As the hydrogen flame propagates within the chamber, the case provides valuable insights into the interaction between turbulent mixing and combustion processes. Such insights are directly relevant to power system applications. For example, understanding the flame stabilization mechanisms and heat release dynamics can inform the design and optimization of hydrogen gas turbines and hybrid power systems. Moreover, the inherent periodicity of the domain further allows for a systematic study of the energy cascade and flame–turbulence interactions, thereby establishing a robust numerical benchmark for power system simulations involving turbulent combustion.

For this case, the CFD simulation task, postprocessing tasks, analysis tasks, and optimization tasks were first defined with natural language as the input.

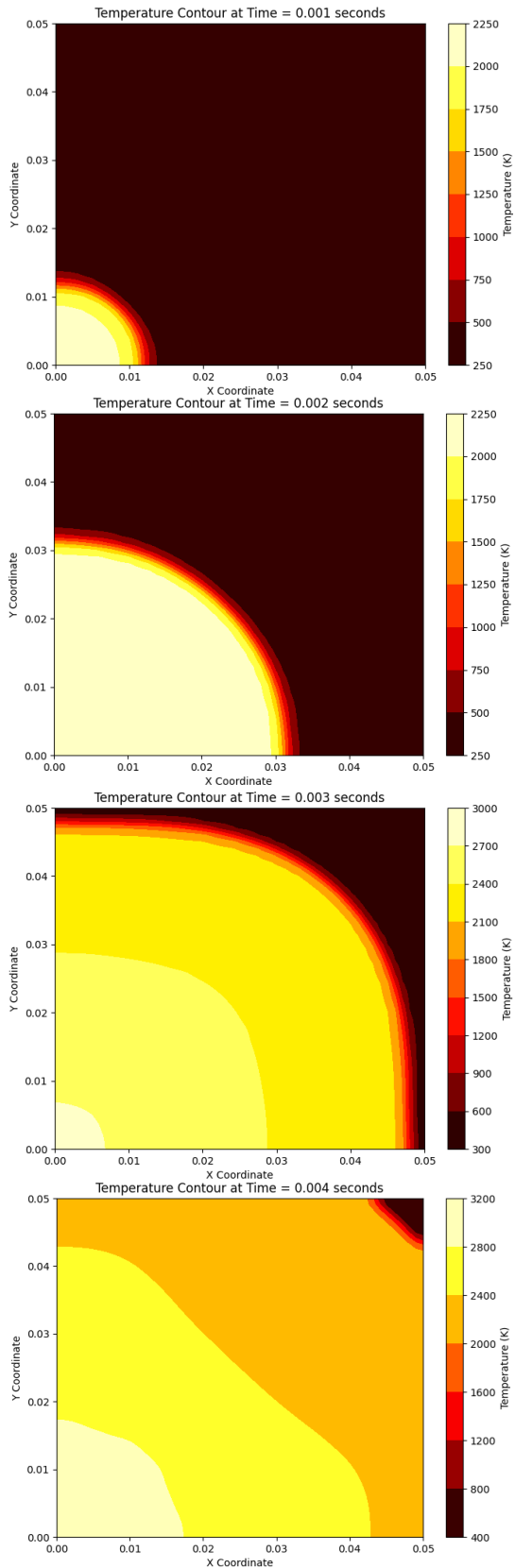

Figure 9: Postprocessing results (contour of temperature) of hydrogen combustion chamber. The images from top to bottom correspond to 1 ms, 2 ms, 3 ms, and 4 ms.

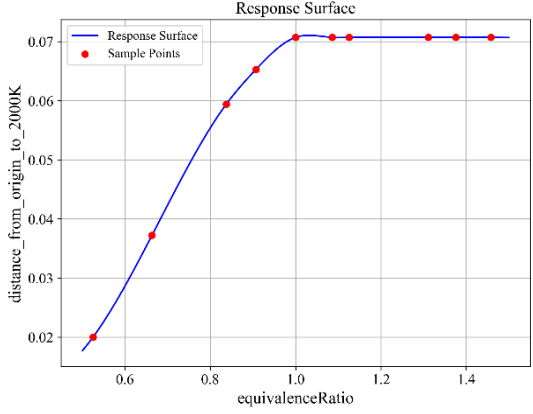

Figure 10: Effect of equivalenceRatio on the distance from the origin in a hydrogen combustion chamber.

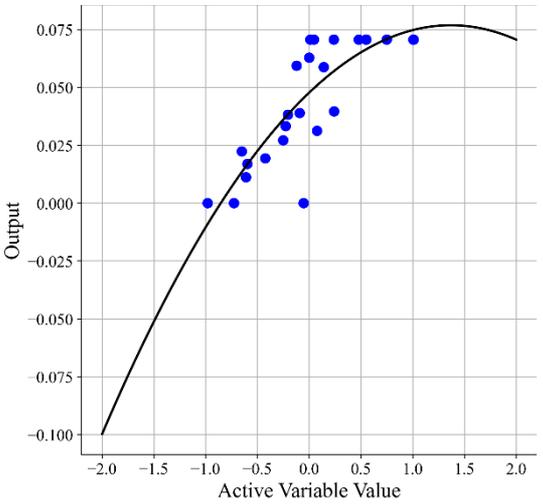

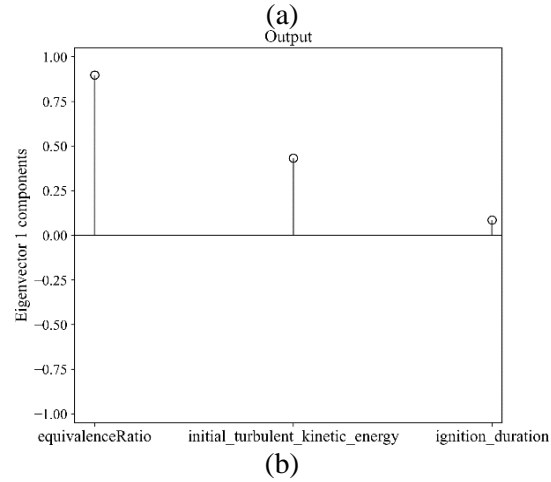

Figure 11: Analysis results of the effect of equivalenceRatio, inital turbulent kinetic energy, ignition duration time on the distance from the origin in a hydrogen combustion chamber simulated by MetaOpenFOAM. (a) response surface build by active subspace method (b) components of active direction ($\hat{\boldsymbol{w}}$).

**CFD simulation task**: Perform a 2D CFD simulation of a hydrogen combustion chamber using a grid size of 50x50x1 with an end time of 0.005 seconds.

**CFD postprocessing tasks**: ① Plot 2D contour plots of temperature at times 0.001, 0.002, 0.003, and 0.004 seconds. ② Extract the locations where the temperature reaches 2000K and compute the corresponding distance from the origin, defined as ($d = sqrt(X^2 + Y^2)$).

**CFD analysis tasks**: ① Analyze the effect of equivalenceRatio (from 0.5 to 1.5) on the distance from the origin, defined as $d = sqrt(X^2 + Y^2)$, where the temperature reaches 2000K at latest time (endTime = 0.005 s) through post-processing (if min(T) > 2000 K, then $d = sqrt(max(X)^2 + max(Y)^2)$; if max(T)<2000 K, then d = 0). ② Analyze the effect of equivalenceRatio (from 0.5 to 1.5), inital turbulent kinetic energy (from 1 to 10), initial ignition duration time (from 0 to 0.002) on the distance from the origin, defined as $d = sqrt(X^2 + Y^2)$, where the temperature reaches 2000K at latest time (endTime = 0.0025 s) through post-processing (if min(T) > 2000 K, then $d = sqrt(max(X)^2 + max(Y)^2)$; if max(T)<2000 K, then d = 0)

**CFD optimization tasks**: Determine the min equivalenceRatio at which d is near d_max (0.0707) at latest time.

Figure 9 shows the visualization results returned by MetaOpenFOAM after executing CFD postprocessing task ①. It is evident that following ignition initiated in the lower left corner, the flame propagates within the hydrogen combustion chamber and nearly covers the entire domain by 4 ms. Figure 10 further illustrates the relationship between the equivalence ratio and the flame propagation radius at 5 ms, as obtained from CFD analysis task ① executed by OptMetaOpenFOAM. The data indicate that as the equivalence ratio increases from 0.5 to 1.5, the flame propagation radius also increases. This behavior arises because the hydrogen flame propagation speed initially increases with the equivalence ratio (when the ratio exceeds 1) before eventually decaying at a slower rate. Moreover, the response surface derived from this analysis allows the determination of the minimum equivalence ratio at which the propagation distance d is nearly maximized, leading to a final optimized value of 1.09. In addition, Figure 11 presents the response surface and the components of $\hat{\boldsymbol{w}}$ obtained by combining CFD analysis task ② executed by OptMetaOpenFOAM with the active subspace method. Within this subspace, the variable d is confined between 0 and 0.0707, corresponding to conditions where the flame is either unpropagated or fully propagated throughout the chamber. Owing to these inherent bounds, the fitted response surface is not perfect. Ultimately, the influence of the three key variables on the flame propagation speed is ranked as follows: equivalence ratio > initial turbulent kinetic energy > ignition duration time. Remarkably, using only five concise prompts of approximately 200 characters each, the entire process—traditionally requiring over 2,000 lines of code for basic visualization, factor analysis, and key parameter optimization—was efficiently executed.

## 5 Conclusion

This study introduces OptMetaOpenFOAM, a pioneering framework that integrates LLM-driven COT methodologies with CFD simulations, enabling automated sensitivity analyses and parameter optimizations. Through natural language inputs, OptMetaOpenFOAM effectively bridges the gap between users and complex CFD workflows, empowering non-experts to execute intricate analyses and optimizations with ease. The integration of external analysis tools, such as the active subspace method and L-BFGS-B optimization algorithm, further enhances the framework's capacity to perform detailed sensitivity analysis and multivariable optimization.

The framework's performance was validated across a diverse range of test cases, including flow, heat transfer, and combustion scenarios. Specifically, a series of tests were conducted on multiple case studies that involved tasks such as univariate optimization and multivariable sensitivity analysis. Results confirm that OptMetaOpenFOAM can accurately interpret user requirements expressed in natural language, effectively set up and execute CFD simulations, and seamlessly integrate external tool libraries with MetaOpenFOAM to complete the tasks.

Notably, the successful validation using a non-OpenFOAM-tutorial hydrogen combustion chamber case demonstrated its efficiency in handling complex combustion dynamics and optimization tasks, where a concise 200-character input triggered comprehensive simulations, analyses, and optimizations, generating over 2,000 lines of code. These findings underline the transformative potential of LLM-driven frameworks in revolutionizing CFD simulation workflows, making them more accessible, efficient, and effective for both industrial and research applications.

## Acknowledgments

This work was supported by the National Natural Science Foundation of China (No. 52025062 and 52106166). The authors also acknowledge High-Performance Computing Centre at Tsinghua University for providing computational resource. During the preparation of this work the author(s) used ChatGPT in order to improve language and readability. After using this tool/service, the author(s) reviewed and edited the content as needed and take(s) full responsibility for the content of the publication.## References

[1] J. Kaplan, S. McCandlish, T. Henighan, T.B. Brown, B. Chess, R. Child, S. Gray, A. Radford, J. Wu, D. Amodei, Scaling laws for neural language models, arXiv preprint arXiv:2001.08361, (2020).
[2] OpenAI, Hello gpt-4o. https://openai.com/index/hello-gpt-4o/.
[3] M. AI, Introducing Llama 3.1: Our most capable models to date. https://ai.meta.com/blog/meta-llama-3-1/.
[4] OpenAI, Introducing OpenAI o1. https://openai.com/o1/.
[5] T. Kumar, Z. Ankner, B.F. Spector, B. Bordelon, N. Muennighoff, M. Paul, C. Pehlevan, C. Ré, A. Raghunathan, Scaling laws for precision, arXiv preprint arXiv:2411.04330, (2024).
[6] DeepSeek, Introducing DeepSeek-V3. https://api-docs.deepseek.com/news/news1226.
[7] D. Guo, D. Yang, H. Zhang, J. Song, R. Zhang, R. Xu, Q. Zhu, S. Ma, P. Wang, X. Bi, Deepseek-r1: Incentivizing reasoning capability in llms via reinforcement learning, arXiv preprint arXiv:2501.12948, (2025).
[8] J. Wei, X. Wang, D. Schuurmans, M. Bosma, F. Xia, E. Chi, Q.V. Le, D. Zhou, Chain-of-thought prompting elicits reasoning in large language models, Adv. Neural Inf. Process. Syst. 35 (2022) 24824-24837.
[9] H. Jasak, A. Jemcov, Z. Tukovic. OpenFOAM: A C++ library for complex physics simulations. In: International workshop on coupled methods in numerical dynamics; 2007: Dubrovnik, Croatia). p. 1-20.

[10] U. Manual, ANSYS FLUENT 12.0, Theory Guide 67 (2009).

[11] C. Multiphysics, Introduction to comsol multiphysics®, COMSOL Multiphysics, Burlington, MA, accessed Feb 9 (1998) 32.

[12] Y. Chen, X. Zhu, H. Zhou, Z. Ren, MetaOpenFOAM: an LLM-based multi-agent framework for CFD, arXiv preprint arXiv:2407.21320, (2024).

[13] Y. Chen, X. Zhu, H. Zhou, Z. Ren, MetaOpenFOAM 2.0: Large Language Model Driven Chain of Thought for Automating CFD Simulation and Post-Processing, arXiv preprint arXiv:2502.00498, (2025).

[14] J. Blazek, Computational fluid dynamics: principles and applications, Butterworth-Heinemann (2015).

[15] J. An, H. Wang, B. Liu, K.H. Luo, F. Qin, G.Q. He, A deep learning framework for hydrogen-fueled turbulent combustion simulation, Int. J. Hydrogen Energ. 45 (2020) 17992-18000.

[16] R. Mao, M. Lin, Y. Zhang, T. Zhang, Z.-Q.J. Xu, Z.X. Chen, DeepFlame: A deep learning empowered open-source platform for reacting flow simulations, Comput. Phys. Commun. 291 (2023) 108842.

[17] Y. Wang, P. Chatterjee, J.L. de Ris, Large eddy simulation of fire plumes, P. Combust. Inst. 33 (2011) 2473-2480.

[18] J. Wei, J. An, N. Wang, J. Zhang, Z. Ren, Velocity nonuniformity and wall heat loss coupling effect on supersonic mixing layer flames, Aerosp. Sci. Technol. 141 (2023) 108545.

[19] A. Badagabettu, S.S. Yarlagadda, A.B. Farimani, Query2cad: Generating cad models using natural language queries, arXiv preprint arXiv:2406.00144, (2024).

[20] J. Weiss. A tutorial on the proper orthogonal decomposition. In: AIAA aviation 2019 forum; 2019. p. 3333.

[21] P.J. Schmid, Dynamic mode decomposition and its variants, Annu. Rev. Fluid Mech. 54 (2022) 225-254.

[22] R. Shan, C.S. Yoo, J.H. Chen, T. Lu, Computational diagnostics for n-heptane flames with chemical explosive mode analysis, Combusti. Flame 159 (2012) 3119-3127.

[23] D. Zhou, N. Schärli, L. Hou, J. Wei, N. Scales, X. Wang, D. Schuurmans, C. Cui, O. Bousquet, Q. Le, Least-to-most prompting enables complex reasoning in large language models, arXiv preprint arXiv:2205.10625, (2022).

[24] D. Paul, R. West, A. Bosselut, B. Faltings, Making Reasoning Matter: Measuring and Improving Faithfulness of Chain-of-Thought Reasoning, arXiv preprint arXiv:2402.13950, (2024).

[25] S. Diao, P. Wang, Y. Lin, R. Pan, X. Liu, T. Zhang, Active prompting with chain-of-thought for large language models, arXiv preprint arXiv:2302.12246, (2023).

[26] X. Wang, J. Wei, D. Schuurmans, Q. Le, E. Chi, S. Narang, A. Chowdhery, D. Zhou, Self-consistency improves chain of thought reasoning in language models, arXiv preprint arXiv:2203.11171, (2022).

[27] P.G. Constantine, E. Dow, Q. Wang, Active subspace methods in theory and practice: applications to kriging surfaces, SIAM J. Sci. Comput. 36 (2014) A1500-A1524.

[28] T.W. Lukaczyk, P. Constantine, F. Palacios, J.J. Alonso. Active subspaces for shape optimization. In: 10th AIAA multidisciplinary design optimization conference; 2014. p. 1171.

[29] D.R.S. Saputro, P. Widyaningsih. Limited memory Broyden-Fletcher-Goldfarb-Shanno (L-BFGS) method for the parameter estimation on geographically weighted ordinal logistic regression model (GWOLR). In: AIP conference proceedings; 2017: AIP Publishing. p.

[30] J.D. Head, M.C. Zerner, A Broyden—Fletcher—Goldfarb—Shanno optimization procedure for molecular geometries, Chem. Phys. Lett. 122 (1985) 264-270.

[31] S. Hong, X. Zheng, J. Chen, Y. Cheng, J. Wang, C. Zhang, Z. Wang, S.K.S. Yau, Z. Lin, L. Zhou, Metagpt: Meta programming for multi-agent collaborative framework, arXiv preprint arXiv:2308.00352, (2023).

[32] H. Chase, LangChain, https://github.com/langchain-ai/langchain.

[33] M. Douze, A. Guzhva, C. Deng, J. Johnson, G. Szilvasy, P.-E. Mazaré, M. Lomeli, L. Hosseini, H. Jégou, The faiss library, arXiv preprint arXiv:2401.08281, (2024).

[34] M. Chen, J. Tworek, H. Jun, Q. Yuan, H.P.D.O. Pinto, J. Kaplan, H. Edwards, Y. Burda, N. Joseph, G. Brockman, Evaluating large language models trained on code, arXiv preprint arXiv:2107.03374, (2021).

[35] Y. Dong, X. Jiang, Z. Jin, G. Li, Self-collaboration code generation via chatgpt, ACM Trans. Software Eng. Method. 33 (2024) 1-38.

[36] A. Mathur, S. He, Performance and implementation of the Launder–Sharma low-Reynolds number turbulence model, Comput. Fluids 79 (2013) 134-139.

## Appendices A. Image Analysis Prompts

The following prompts are used for image analysis in the context of studying the relationship between independent and dependent variables using the Active Subspace Method.

*Prompt for Multi-Variable Image Analysis*

PROMPT_for_image_analyze_multi_vars: str = """
  *These two images analyze the relationship between the independent variables (independent_vars) and the dependent variable (dependent_var) using the Active Subspace Method. The first figure presents the response surface constructed by the Active Subspace Method, while the second figure illustrates the magnitude of influence of each parameter.*
  *The user's requirement is user_requirements.*
  *Please analyze two images and respond to the user's requirement accordingly.*
  """

*Prompt for Single-Variable Image Analysis*

PROMPT_for_image_analyze: str = """
  *This image illustrates the trend between independent_var and dependent_var. The user's requirement is user_requirements.*
  *Please analyze this image and respond to the user's requirement accordingly.*
  """

These prompts are designed to guide the analysis of images based on user requirements, focusing on understanding variable relationships and parameter influences in computational models.

## Appendices B. Textual Response for the Analysis Task

### B.1 PitzDaily

*Univariate Analysis*

Analysis of the Effect of Inlet Flow Velocity on Max y+ in a RANS Simulation

- Interpretation of the Provided Image

The image presents a response surface plot showing the relationship between **inlet flow velocity** (x-axis) and **max y+** (y-axis). Key observations include:

- A **positive linear trend**, where an increase in inlet flow velocity results in an increase in max y+.
- Sample points (red dots) align well with the response surface (blue line), indicating a good model fit.
- The range of **inlet flow velocity** appears to be between **8.0 and 12.0**.
- The corresponding **max y+** values increase from approximately **22 to 31**.

- Effect of Inlet Flow Velocity on Max y+
1. **Linear Relationship**: The results suggest a nearly **linear increase** in max y+ as inlet flow velocity increases. This aligns with theoretical expectations, as increasing velocity leads to higher shear stress at the wall, increasing y+.
2. **Turbulent Boundary Layer Influence**: Since the max y+ values increase, it suggests that with higher inlet velocities, the boundary layer becomes thinner, and the near-wall resolution requirement increases.
3. Implications for Mesh Resolution:
    - For **low velocities (~8 m/s)**, the max y+ is about 22, which might still be within the **low-Reynolds number** region.
    - For **higher velocities (~12 m/s)**, max y+ reaches **over 30**, possibly pushing into the logarithmic layer. This suggests that **wall function approaches** might be necessary for accurate turbulence modeling.

Recommendations for Your RANS Simulation

1. Mesh Refinement Strategy:
    - If you aim for a **low-Re** treatment (e.g., y+ < 1), further **refining the first cell height** is required.
    - If using wall functions, a y+ range of 30-100 is acceptable.
    - Your max y+ values (22-31) suggest that you are close to **transitioning between these regimes**. Ensure that the mesh resolution is adequate for the turbulence model you use.
2. Turbulence Model Considerations:
    - **k−ωk-\omegak−ω SST**: If using a low-y+ approach, this model is suitable, but mesh adaptation might be needed.
    - **Standard k−ϵk-\epsilonk−ϵ**: Works well with wall functions, ensuring that y+ stays in the 30+ range.
3. Simulation Setup for pitzDaily Using pimpleFoam:
    - Run a **steady-state RANS simulation** using pimpleFoam (though typically used for transient cases).
    - Set up **boundary conditions** appropriately for different velocity values.
    - Ensure **convergence criteria** are met for each case to confirm the trend.

Would you like me to assist in setting up the OpenFOAM case files or optimizing the mesh for your target y+ values?

*Multivariable Analysis*

Analysis of the Effect of Inlet Flow Velocity and

Inlet Turbulent Kinetic Energy on max y+

1. Understanding the Response Surface (First Image)
- The response surface in the first figure visualizes the relationship between an **active variable** (a combination of the independent parameters) and the dependent variable **max y+**.
- The black line represents the model's predicted output, while the blue dots show the actual simulation results.
- The strong linear trend in the response surface suggests that max y+ is **linearly dependent** on the active variable.
- This indicates that the parameters used in constructing the active variable have a dominant influence on max y+ +.

2. Influence of Parameters (Second Image)
- The second figure presents the eigenvector components, which indicate the contribution of **inlet flow velocity** and **inlet turbulent kinetic energy** to the active subspace.
- The **inlet flow velocity** has an eigenvector component close to **1.0**, whereas the **inlet turbulent kinetic energy** has an eigenvector component close to **0.0**.
- This suggests that inlet flow velocity is the dominant factor affecting max y+, while inlet turbulent kinetic energy has a negligible effect.

---

- Recommendation for RANS Simulation (pitzDaily Case Using pimpleFoam)

To further analyze the effect of inlet flow velocity and inlet turbulent kinetic energy on max y+ you should:

1. Set Up the pitzDaily Case
    - Use OpenFOAM's standard **pitzDaily** case (backward-facing step flow) as the geometry.
    - Configure the simulation to run with **incompressible flow** using the pimpleFoam solver.
2. Modify the Boundary Conditions
    - **Vary the inlet flow velocity** systematically (e.g., test multiple values: 5 m/s, 10 m/s, 15 m/s).
    - **Vary the inlet turbulent kinetic energy** (though the analysis suggests it has little impact, it's still useful to confirm).
3. Monitor max y+ Values
    - Run the simulation for different conditions and extract **wall y+ values** at critical locations (especially near the separation and reattachment regions).
    - Compare the results to confirm whether inlet flow velocity is indeed the primary driver.
4. Post-Processing & Validation
    - Use postProcess -func yPlus in OpenFOAM to obtain max y+.
    - Plot the results and check consistency with the Active Subspace analysis.

---

- Conclusion
  - Based on the response surface and eigenvector analysis, inlet flow velocity significantly affects max y+, while inlet turbulent kinetic energy has minimal impact.
  - Your RANS simulation in OpenFOAM should focus on varying **inlet flow velocity** to validate this observation.
  - Use **pimpleFoam** for time-accurate results and ensure appropriate wall functions are applied to capture the boundary layer effects correctly.

Would you like help setting up the OpenFOAM simulation scripts? 🚀

## B.2 HIT

*Univariate Analysis*

**Analysis of the Effect of Laminar Viscosity (ν) on Turbulent Kinetic Energy**

1. **Trend Observation**:
    - The **response surface (blue curve)** represents the variation of **average turbulent kinetic energy** with respect to **laminar viscosity**.
    - The **sample points (red dots)** show the discrete data used to construct the response surface.
    - The relationship is **nonlinear** and exhibits a **strong inverse correlation**.
2. **Key Findings**:
    - As **laminar viscosity (ν) increases** (from 0.01 to 0.1), the **average turbulent kinetic energy decreases significantly**.
    - For **low viscosity values (ν≈0.01)**, the kinetic energy is **relatively high** (~0.02).
    - As ν increases towards 0.1, the turbulent kinetic energy **drops rapidly** and approaches nearly **zero**.
    - The trend suggests that increasing viscosity **dampens turbulence intensity** by dissipating energy more effectively.
3. **Physical Explanation**:
    - In DNS simulations of turbulence,

**viscosity (ν) governs the dissipation rate of kinetic energy**.
  - Higher viscosity leads to **stronger energy dissipation**, reducing the turbulence intensity.
  - At low viscosity, turbulent eddies **persist longer**, leading to higher **average kinetic energy**.
  - At high viscosity, the turbulence **decays rapidly**, suppressing kinetic energy.

**Recommendations for DNS Simulation Setup (dnsFoam, boxTurb, Grid 16^3)**
1. **Case Setup**:
   - Ensure **correct Reynolds number scaling** by selecting appropriate **forcing terms**.
   - Choose **sufficient time steps** to capture turbulence dynamics accurately.
   - Verify the **Kolmogorov scales** are resolved within the given **$163 16^3 163$ grid**.
2. **Validation**:
   - Compare simulation results with theoretical **energy spectra** (e.g., Kolmogorov's $k^{(-5/3)}$ scaling in the inertial subrange).
   - Ensure proper statistical convergence by running the simulation for a sufficiently long time.
3. **Additional Considerations**:
   - If higher resolution is needed, consider increasing the grid size (e.g., $32^3$ or $64^3$).
   - Use **energy spectra analysis** to confirm the dissipation range behavior at different viscosity values.

Would you like assistance in setting up the **dnsFoam case** or post-processing the turbulence statistics?

**B.3 BuoyantCavity**

*Univariate Analysis*

Analysis of the Effect of Temperature Difference on Max Velocity in X Direction
- Overview of the Simulation:

The simulation conducted is a RANS (Reynolds-Averaged Navier-Stokes) simulation of buoyantCavity using buoyantFoam. This setup is designed to analyze natural convection in a heated cavity, where:
- **Hot and cold patches** act as temperature boundaries.
- Remaining patches are treated as adiabatic (no heat flux).
- **Buoyancy effects** drive the flow, leading to natural convection.

- Analysis of the Response Surface:

The provided response surface plot illustrates the relationship between temperature difference (ΔT) and maximum velocity in the X direction (Umax,x).
1. Observations from the Plot:
   - The response surface follows a smooth, **nonlinear** increasing trend.
   - As ΔT increases from 10 K to 30 K, Umax,x also increases.
   - The **growth is nonlinear**, indicating an accelerating effect of temperature difference on the velocity.
   - Sample points (red dots) confirm the trend predicted by the response surface (blue curve).
2. Physical Interpretation:
   - Natural convection is driven by **buoyancy forces** induced by the temperature gradient.
   - Higher **temperature difference (ΔT)** enhances buoyancy, leading to stronger convection currents.
   - The acceleration in **Umax,x** suggests that at higher ΔT, flow becomes more vigorous, possibly leading to increased turbulence effects.
3. Key Takeaways:
   - The relationship between ΔT and Umax,x is **not linear**—it suggests **quadratic or exponential growth**.
   - At **lower ΔT**, natural convection is relatively weak, leading to slower velocities.
   - At **higher ΔT**, natural convection dominates, significantly increasing the max velocity.

- Suggested Further Analysis:

To gain deeper insight into the natural convection behavior in the buoyant cavity, consider the following:
1. Flow Pattern Analysis:
   - Visualize velocity contours and streamlines to understand the flow structure.
   - Identify potential vortex formations and their intensity changes with ΔT.
2. Turbulence Effects:
   - Investigate the transition from laminar to turbulent convection at higher ΔT.
   - Analyze turbulence kinetic energy

(TKE) and Reynolds stresses.
3. Heat Transfer Performance:
    o Compute Nusselt number (Nu) at the hot and cold walls to quantify convective heat transfer enhancement.
    o Compare with theoretical correlations for natural convection in cavities.

- Conclusion:

The results indicate that as temperature difference increases, the buoyancy-driven flow becomes significantly stronger, leading to a nonlinear rise in maximum velocity in the X direction. This highlights the importance of ΔT in controlling natural convection intensity in enclosures.

Would you like further assistance in extracting numerical trends or performing additional post-processing on your data?

*Multivariable Analysis*

Analysis of the Effect of Independent Variables on Max Velocity in X Direction

- Overview

The analysis utilizes the Active Subspace Method (ASM) to investigate how four independent parameters affect the maximum velocity in the X direction in a RANS (Reynolds-Averaged Navier-Stokes) simulation of a buoyant cavity using the buoyantFoam solver and the kEpsilon turbulence model.

- Interpretation of the First Figure: Response Surface
- The first figure presents a **response surface**, constructed using ASM, which models the relationship between an **active variable** (a linear combination of the four parameters) and the **dependent variable (max velocity in X direction)**.
- The **blue points** represent the actual simulation results, while the **black curve** is the response surface approximation.
- The **trend** suggests that the velocity initially increases with the active variable and then plateaus or slightly decreases at higher values, indicating a nonlinear but predictable relationship.
- Interpretation of the Second Figure: Eigenvector Contributions
- The second figure presents the **magnitude of influence of each independent variable** in the first eigenvector (dominant active subspace direction).
- The **temperature difference between hot and cold surfaces** has the highest magnitude, meaning it has the strongest impact on the velocity.
- The **other three parameters (k, epsilon, Prt)** have significantly smaller magnitudes, indicating they contribute much less to variations in the max velocity.
- Conclusions and Insights
1. Dominance of Temperature Difference:
    o Since **temperature_difference_between_hot_and_cold** has the largest eigenvector component (close to 1.0), it is the **primary driver** of max velocity changes in the cavity.
    o Increasing the temperature difference (from 10 K to 30 K) will likely result in a significant increase in velocity due to enhanced buoyancy forces driving the convection.
2. Minor Influence of k, epsilon, and Prt:
    o The parameters k_of_all_boundarys, epsilon_of_all_boundarys, and Prt_of_all_boundarys_in_alphat have negligible impact on velocity.
    o This suggests that **turbulence properties do not strongly influence** the primary flow dynamics, possibly due to **natural convection dominance** rather than turbulence-driven effects.
3. Implications for Simulation Setup:
    o If the objective is to **control or optimize the velocity**, adjusting the **temperature difference** is the most effective approach.
    o The **turbulence parameters (k, epsilon, Prt)** should still be chosen appropriately for numerical stability but are **not critical factors** in determining max velocity.
    o Future studies might explore a **larger range** of turbulence parameter variations to confirm their secondary influence.
- Recommendation for Further Investigation
- To validate the findings, consider **running additional RANS simulations** focusing on temperature variation while keeping turbulence parameters fixed.
- Investigate the effect of **even higher temperature differences** (e.g., beyond 30 K) to determine if velocity follows the same nonlinear trend observed in the response surface.
- Consider performing **LES (Large Eddy Simulation) studies** to verify whether turbulence plays a more significant role at higher Rayleigh numbers.

Would you like further numerical analysis or additional visualizations for better insight?

**B.4 CounterFlowFlame2D**

*Univariate Analysis*

1. Analysis of the Effect of Inlet Velocity on Maximum Temperature
2. Observations from the Image

The graph illustrates the relationship between **inlet velocity** (x-axis) and **maximum temperature** (y-axis) in a **counterflow flame** simulation. The key observations are:

3. **Inverse Relationship**: As inlet velocity increases from **10 m/s to 60 m/s**, the maximum temperature **decreases** significantly.
4. **High Temperature at Low Velocities**: The maximum temperature is highest (~1800 K) at the lowest inlet velocity.
5. **Gradual Decline**: Up to around **35-40 m/s**, the maximum temperature decreases gradually.
6. **Steep Drop Beyond 40 m/s**: A sharp decline is observed beyond **40 m/s**, where the maximum temperature drops from ~1200 K to below 400 K.
7. **Near Constant Temperature at Higher Velocities**: Beyond **50 m/s**, the temperature remains nearly constant at a low value.
8. Possible Physical Interpretation
- **Flame Stabilization**: At low velocities, the flame remains stable, allowing for higher temperatures.
- **Stronger Heat Dissipation at High Velocities**: Increased velocity leads to more **convective cooling**, reducing the peak temperature.
- **Extinction Limit**: The sharp drop after **40 m/s** suggests a **flame extinction limit**, where the flame cannot sustain itself due to excessive strain and turbulence effects.
9. Implications for the Simulation
- **Choice of Inlet Velocity**: If the goal is to study **stable combustion**, inlet velocity should be kept below **40 m/s**.
- Flame Quenching: The sharp drop in temperature beyond 40 m/s should be further analyzed to identify whether it corresponds to strain-induced flame quenching.
- Further Simulations: Additional 2D laminar reactingFoam simulations should focus on:
  o Confirming critical velocity for extinction.
  o Analyzing **species concentration** and **reaction rates** at different velocities.
  o Investigating **flame structure** for varying velocities.

Would you like help setting up a **reactingFoam simulation** for this case?

*Multivariable Analysis*

The two images you provided analyze the relationship between the independent variables (inlet_velocity and inlet_temperature) and the dependent variable (max_temperature) in a laminar simulation using the Active Subspace Method.

- Image 1: Response Surface

The first image represents the response surface generated by the Active Subspace Method. Here's a breakdown:

- The **x-axis** represents the **inlet velocity**, ranging from -2.0 to 2.0, corresponding to a range of velocities from 10.0 to 60.0 m/s (given your request).
- The **y-axis** represents the **inlet temperature**, which ranges from approximately 0 to 2.0, corresponding to 243 K to 343 K.
- The **black curve** shows the underlying model response, while the **blue points** correspond to sampled data points.

From the figure, it can be observed that as **inlet velocity** increases, the **max_temperature** decreases (following the negative slope of the curve), indicating that higher inlet velocity tends to reduce the maximum temperature. Similarly, the inlet temperature has a notable impact on the **max_temperature**, with higher temperatures resulting in an increase in the max temperature. The response surface is concave and downward sloping for increasing inlet velocity and inlet temperature.

- Image 2: Eigenvector 1 Components

The second image shows the contributions of the two independent variables (inlet_velocity and inlet_temperature) to the first eigenvector of the system:

- The **inlet velocity** shows a very high component along the first eigenvector (near 1.0), suggesting that it has a dominant influence on the max temperature.
- The **inlet temperature** shows a much smaller component (close to 0), indicating it has a negligible effect when compared to inlet velocity in determining the max temperature.
- Conclusion

Based on the analysis of both images:

1. **Inlet velocity** has a dominant effect on max temperature, with a strong contribution to the first eigenvector (near 1.0). This suggests that inlet velocity is a major factor controlling the behavior of the max temperature in the simulation. As inlet velocity increases, max temperature tends to decrease.
2. **Inlet temperature**, on the other hand, appears to have a minor effect, as indicated by its small eigenvector component (close to 0).

So, if you're conducting a simulation in a counterflow flame scenario, focusing on controlling **inlet velocity** will have a significant impact on **max_temperature**, while **inlet temperature** will have a lesser influence in comparison.

### B.5 Hydrogen Combustion Chamber

*Univariate Analysis*

The image depicts the relationship between the equivalence ratio (ranging from 0.5 to 1.5) and the distance from the origin at which the temperature reaches 2000K in a hydrogen combustion chamber simulation.

**Analysis:**

1. **Trend Overview:**
   - The plot indicates that as the equivalence ratio increases from **0.5 to around 1.0**, the distance **steadily increases**.
   - Beyond an equivalence ratio of approximately **1.0**, the distance **saturates** at around **0.07**.

2. **Physical Interpretation:**
   - At **low equivalence ratios (lean conditions, φ < 1.0)**, combustion is less intense, and the flame is located closer to the origin.
   - As the equivalence ratio increases towards **stoichiometric conditions (φ ≈ 1.0)**, combustion becomes more vigorous, and the flame stabilizes further from the origin.
   - For **rich conditions (φ > 1.0)**, the flame position does not shift significantly, suggesting that either:
     - The mixture is fuel-rich, limiting additional reaction intensity.
     - The temperature distribution stabilizes due to the combustion process reaching a peak efficiency.

3. **Simulation & Post-Processing Considerations:**
   - The methodology used to compute **d = sqrt (X² + Y²)** ensures that if the **minimum temperature is above 2000K**, the distance corresponds to the farthest grid point in the simulation.
   - If **max(T) < 2000K**, d = 0, indicating no significant combustion event producing high temperatures in the chamber.

**Conclusion:**

- The **distance from the origin where T = 2000K increases with equivalence ratio up to φ ≈ 1.0**, beyond which it **plateaus**.
- This suggests that an **optimal range for effective flame propagation in the given combustion chamber exists around φ = 1.0**.
- Beyond this, the system does not exhibit significant changes, indicating that increasing fuel content does not push the reaction zone further.

Would you like a deeper discussion on optimizing combustion performance based on these results?

*Multivariable Analysis*

**Analysis of the Active Subspace Method Results**
The two images provided present an analysis of the relationship between the independent variables:
- **Equivalence Ratio** (ranging from 0.5 to 1.5)
- **Initial Turbulent Kinetic Energy** (ranging from 1 to 10)
- **Ignition Duration** (ranging from 0 to 0.002)

and their effect on the dependent variable:
- **Distance from the Origin where Temperature Reaches 2000K**.

**Image 1: Response Surface**
- The first figure shows a **response surface** constructed using the **Active Subspace Method**.
- The **x-axis** represents the **active variable** obtained through the method, which is a linear combination of the independent variables.
- The **y-axis** represents the output variable (**distance from origin where temperature reaches 2000K**).
- The blue scatter points represent **simulation data**, and the black curve is the

**fitted response surface**.
- The general trend suggests a **nonlinear correlation**, with distance increasing initially and then leveling off.

**Image 2: Sensitivity of Parameters**
- The second figure presents the **magnitude of influence** of each independent variable in defining the **active variable** (Eigenvector 1 components).
- The bars indicate the weight (importance) of each variable:
  - **Equivalence Ratio** has the highest influence (~0.9).
  - **Initial Turbulent Kinetic Energy** has a moderate influence (~0.5).
  - **Ignition Duration** has a minor influence.

**Implications for the Simulation Setup**
Based on these findings:
1. **Equivalence Ratio** is the dominant factor affecting the **distance from the origin where temperature reaches 2000K**. The simulations should focus on its variations from **0.5 to 1.5**.
2. **Initial Turbulent Kinetic Energy** also plays a role, though not as significant. Testing values from **1 to 10** will provide additional insight into its effect.
3. **Ignition Duration** has a relatively low impact, but for completeness, it should still be varied from **0 to 0.002**.

**Simulation Setup for 2D Hydrogen Combustion Chamber**
Given the user's requirements:
- **Grid:** 50×50×150 \times 50 \times 150×50×1
- **End Time:** 0.0050.0050.005 seconds
- **Combustion Process:** Hydrogen-air reaction

**Post-Processing Criterion**
To determine the **distance from the origin (d)** where temperature reaches **2000K**, the following logic applies:
- If the **minimum temperature (T)** in the domain is **greater than 2000K**, d = d_max
- If the **maximum temperature (T)** is **less than 2000K**, then: d=0

**Next Steps**
1. **Run the 2D simulation** of the hydrogen combustion chamber for various combinations of:
   - Equivalence Ratio (**0.5 to 1.5**)
   - Initial Turbulent Kinetic Energy (**1 to 10**)
   - Ignition Duration (**0 to 0.002**)
2. **Post-process the simulation results** to extract the **distance d** using the criteria above.
3. **Compare the simulation results** to the **Active Subspace Method's response surface** and validate the accuracy.

Would you like assistance in setting up the simulation framework or interpreting the results after running the simulation?

# Appendices C. Detailed Thought and Response

The following displays all the thought processes as well as part of the output：

## ▽ PitzDaily Parameter Analysis

*Please help me analyze the effect of the inlet flow velocity on max yplus in a simulation: do a RANS simulation of incompressible pitzDaily flow using pimpleFoam*

<think>
**Prepare**
The user's requirement can be divided into the following tasks:
    "CFD_simulation_task": "do a RANS simulation of incompressible pitzDaily flow using pimpleFoam",
    "CFD_postprocessing_task": "extract max yplus at latest time through post-processing",
    "CFD_analysis_task": "analyze the effect of the inlet flow velocity on extract max yplus",
    "CFD_optimization_task": "None"
A similar case is matched:
    " case name": " pitzDaily", " case domain ": " incompressible ",
    " case category ": " RAS", " case solver ": " pimpleFoam "
The independent and dependent variables are:
    "inlet_flow_velocity" and "max_yplus"
**Sampling**
Since the user's request did not specify a range for the independent variable, a default range was selected around the nominal value of 10.0, with a lower bound of 8.0 and an upper bound of 12.0. Five sample points were chosen within this interval. Multiple CFD simulation and postprocessing tasks were then generated accordingly.
**CFD Simulation Task & CFD Postprocessing Task**
For one of these cases, MetaOpenFOAM 2.0 was executed until it successfully passed LLM verification. The input file for this case was subsequently duplicated, and the value corresponding to the sampling point was modified. OpenFOAM was then used to run the simulation again, repeating the process until all cases had been executed.
**CFD Sensitivity Analysis Task**
The postprocessing results from each case were extracted and combined with the sampling points for function fitting. The analysis revealed that a linear fit provided the best approximation. Finally, the results were presented through both graphical illustrations and textual explanations.
</think>

## ▽ PitzDaily Optimization

*Please help me analyze the effect of the inlet flow velocity and inlet turbulent kinetic energy on max yplus and determine the optimal inlet flow velocity and inlet turbulent kinetic energy at which max yplus is blow 25 in a simulation: do a RANS simulation of incompressible pitzDaily flow using pimpleFoam*

*<think>*
*(Same to the above analysis)*
*\*\*CFD Optimization Task\*\**
*The optimization objective is defined as minimizing (yplus−25)^2. Based on the response surface established in the previous section, the optimization is performed within the inlet flow velocity range of 8 to 12 m/s.*
*</think>*

## ▽ PitzDaily Multivariable Analysis

*Please help me analyze the effect of the inlet flow velocity and inlet turbulent kinetic energy on max yplus in a simulation: do a RANS simulation of incompressible pitzDaily flow using pimpleFoam*

*<think>*
*Similar to the thought process in the univariate analysis, the only modification is to the CFD_analysis_task, which is now defined as:*
*"Analyze the effect of inlet flow velocity and inlet turbulent kinetic energy on max yPlus."*
*And the independent variables change to:*
*    "inlet_flow_velocity" and "inlet_turbulent_kinetic_energy"*
*The sampling space is expanded to two dimensions, with eight sample points selected. After executing the simulations, postprocessing results from each case are extracted and combined with the sampled data for active subspace analysis. The results are then presented through graphical visualizations and textual explanations.*
*</think>*

*<output>*
*• Analysis of the Effect of Inlet Flow Velocity and Inlet Turbulent Kinetic Energy on max y+*
*…*
*</output>*

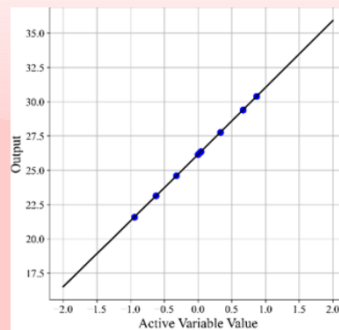
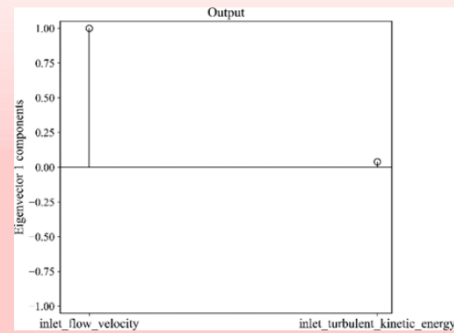

## ▽ HIT Univariate Analysis

*Please help me analyze the effect of the laminar viscosity nu_in_physicalProperties (from 0.01 to 0.1) on the average turbulent kinetic energy (average(1/2*U^2)) in the simulation: do a DNS simulation of incompressible forcing homogeneous isotropic turbulence (boxTurb) using dnsFoam with Grid 16^3*

<think>
**Prepare**
The user's requirement can be divided into the following tasks:
   "CFD_simulation_task": "do a DNS simulation of incompressible forcing homogeneous isotropic turbulence (boxTurb) using dnsFoam with Grid 16^3",
   "CFD_postprocessing_task": "extract the average turbulent kinetic energy (average(1/2*U^2)) at latest time through post-processing",
   "CFD_analysis_task": "analyze the effect of the laminar viscosity nu_in_physicalProperties (from 0.01 to 0.1) on the average turbulent kinetic energy",
   "CFD_optimization_task": "None"
A similar case is matched:
   " case name": " boxTurb16", " case domain ": " DNS",
   " case category ": " None", " case solver ": " dnsFoam"
The independent and dependent variables are:
   "laminar_viscosity_nu_in_physicalProperties" and
   "average_turbulent_kinetic_energy"
**Sampling**
Sampling between 0.01 to 0.1 with 9 points
(Else similar with PitzDaily)
</think>

## ▽ HIT Optimization

*Please help me analyze the effect of the laminar viscosity nu_in_physicalProperties (from 0.01 to 0.1) on the average turbulent kinetic energy (average(1/2*U^2)) and determine the optimal nu_in_physicalProperties at which the average turbulent kinetic energy is near 0.01 in the simulation: do a DNS simulation of incompressible forcing homogeneous isotropic turbulence (boxTurb) using dnsFoam with Grid 16^3*

<think>
(Same to the above analysis)
**CFD Optimization Task**
Similar to the thought process in the previous section, the only modification is to the CFD_optimization_task, which is now defined as:
"Determine the optimal nu in physicalProperties where the average turbulent kinetic energy is close to 0.01."
After completing the CFD_analysis_task, the optimization objective is established as (dependent variable - 0.01)^2. The optimization is then performed within the defined input range to find the optimal value.
</think>

## ▽BuoyantCavity Univariate Analysis

*Please help me analyze the effect of the temperature difference betwwen the hot and cold (from 10 K to 30 K) on the max velocity in X direction in a simulation: do a RANS simulation of buoyantCavity using buoyantFoam, which investigates natural convection in a heat cavity; the remaining patches are treated as adiabatic (except hot and cold patches).*

<think>
**Prepare**
The user's requirement can be divided into the following tasks:
   "CFD_simulation_task": "do a RANS simulation of buoyantCavity using buoyantFoam",
   "CFD_postprocessing_task": "extract the max velocity in X direction at latest time through post-processing",
   "CFD_analysis_task": "analyze the effect of the temperature difference between the hot and cold patches on the max velocity in X direction",
   "CFD_optimization_task": "None"
A similar case is matched:
   " case name": " buoyantCavity ", " case domain ": " DNS",
   " case category ": " None", " case solver ": " buoyantFoam "
The independent and dependent variables are:
   "temperature_difference_between_hot_and_cold_patches" and
   "max_velocity_in_X_direction"
(Else similar with PitzDaily)
</think>

<output>
Analysis of the Effect of Temperature Difference on Max Velocity in X Direction
1. Observations from the Plot:
o  The response surface follows a smooth, nonlinear increasing trend.
o  As $\Delta T$ increases from 10 K to 30 K, $U_{max, x}$ also increases.
o   The growth is nonlinear, indicating an accelerating effect of temperature difference on the velocity.
2. Physical Interpretation:
o   Natural convection is driven by buoyancy forces induced by the temperature gradient....
</output>

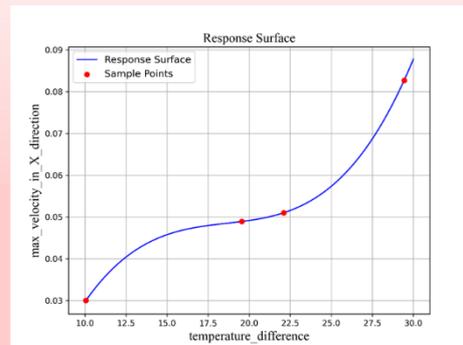

## ▽BuoyantCavity Multivariable Analysis

*Please help me analyze the effect of the temperature difference betwwen hot and cold (from 10 K to 30 K), k_of_all_boundarys (from 1e-04 to 1e-03), epsilon_of_all_boundarys (from 1e-06 to 1e-05) and Prt_of_all_boundarys_in_alphat (from 0.6 to 1.0) on the max velocity in X direction in a simulation: do a RANS simulation of buoyantCavity using buoyantFoam and kEpsilon turbulent model, which investigates natural convection in a heat cavity; the remaining patches are treated as adiabatic (except hot and cold patches).*

<think>
Similar to the thought process in the univariate analysis, the only modification is to the CFD_analysis_task, which is now defined as:
"analyze the effect of temperature_difference_betwwen_hot_and_cold, k_of_all_boundarys, epsilon_of_all_boundarys, and Prt_of_all_boundarys_in_alphat on the max velocity in X direction"
And the independent variables change to:
     "temperature_difference_between_hot_and_cold",
     "k_of_all_boundarys", "epsilon_of_all_boundarys"
     "Prt_of_all_boundarys_in_alphat"
The sampling space is expanded to four dimensions, with 32 sample points selected. After executing the simulations, postprocessing results from each case are extracted and combined with the sampled data for active subspace analysis. The results are then presented through graphical visualizations and textual explanations.
</think>

<output>
• Analysis of the Effect of Independent Variables on Max Velocity in X Direction
...
</output>

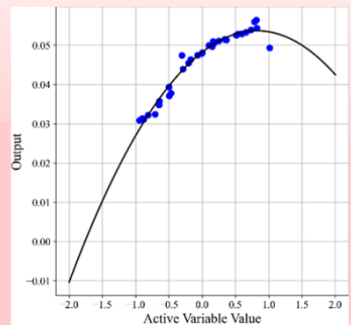
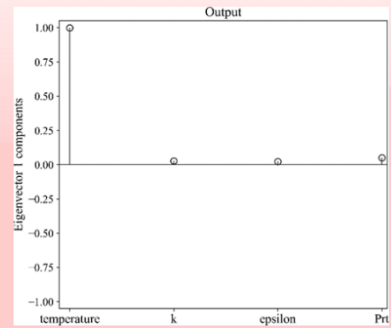

## ▽BuoyantCavity Optimization

*Please help me analyze the effect of the temperature difference betwwen the hot and cold (from 10 K to 30 K) on the max velocity in X direction and determine the optimal temperature difference betwwen the hot and cold at which max velocity in X direction is near 0.07 m/s in a simulation: do a RANS simulation of buoyantCavity using buoyantFoam, which investigates natural convection in a heat cavity; the remaining patches are treated as adiabatic (except hot and cold patches).*

<think>
*(Same to the above analysis)*
**CFD Optimization Task**
Similar to the thought process in the previous section, the only modification is to the CFD_optimization_task, which is now defined as:
"determine the optimal temperature difference betwwen the hot and cold at which max velocity in X direction is near 0.07 m/s in a simulation."
After completing the CFD_analysis_task, the optimization objective is established as (dependent variable - 0.07)^2. The optimization is then performed within the defined input range to find the optimal value.
</think>

<output>
*(Same to the above analysis)*
Optimization of laminar viscosity on average turbulent kinetic energy
• Optimized laminar viscosity: 0.0201
• Minimized objective function value: $3.06 \times 10^{-18}$
</output>

## ▽CounterFlowFlame2D Univariate Analysis

*Please help me analyze the effect of inlet velocity (from 10.0 to 60.0 m/s) on max temperature in the simulation: do a 2D laminar simulation of counterflow flame using reactingFoam in combustion*

<think>
**Prepare**
The user's requirement can be divided into the following tasks:
  "CFD_simulation_task": "do a 2D laminar simulation of counterflow flame using reactingFoam with endTime = 0.1",
  "CFD_postprocessing_task": "extract the max temperature at latest time through post-processing",
  "CFD_analysis_task": "analyze the effect of inlet velocity (from 30.0 to 60.0 m/s) on max temperature",
  "CFD_optimization_task": "None"
A similar case is matched:
  " case name": " buoyantCavity ", " case domain ": " DNS",
  " case category ": " None", " case solver ": " buoyantFoam "
The independent and dependent variables are:
  "inlet_velocity" and "max_temperature"
*(Else similar with PitzDaily)*
</think>

## ▽ CounterFlowFlame2D Multivariable Analysis

Please help me analyze the effect of inlet velocity (from 10.0 to 60.0 m/s) and inlet temperature (from 243 to 343 K) on max temperature in the simulation: do a 2D laminar simulation of counterflow flame using reactingFoam in combustion

<think>
Similar to the thought process in the univariate analysis, the only modification is to the CFD_analysis_task, which is now defined as:
"analyze the effect of inlet velocity and
inlet temperature on max temperature in the simulation"
And the independent variables change to:
"inlet_velocity" and "inlet_temperature"
The sampling space is expanded to two dimensions, with 9 sample points selected.
After executing the simulations, postprocessing results from each case are extracted and combined with the sampled data for active subspace analysis.
The results are then presented through graphical visualizations and textual explanations.
</think>

## ▽ CounterFlowFlame2D Optimization

Please help me analyze the effect of inlet velocity (from 10.0 to 60.0 m/s) on max temperature and determine the optimal inlet velocity at which max temperature is blow 1000 K in the simulation: do a 2D laminar simulation of counterflow flame using reactingFoam in combustion

<think>
(Same to the above analysis)
**CFD Optimization Task**
Similar to the thought process in the previous section, the only modification is to the CFD_optimization_task, which is now defined as:
"determine the optimal inlet velocity at which max temperature is blow 1000 K in the simulation"
After completing the CFD_analysis_task, the optimization objective is established as (dependent variable - 1000)^2. The optimization is then performed within the defined input range to find the optimal value.
</think>

<output>
(Same to the above analysis)
Optimization of inlet velocity on max temperature
• Optimized inlet velocity: 41.2 m/s
• Minimized objective function value: $7.28 \times 10^{-9}$
</output>